\documentclass[conference]{IEEEtran}
\usepackage{amsmath,amssymb,graphicx,epstopdf,cite,subfigure}
\usepackage{psfrag}
\usepackage{amssymb}
\usepackage{amsmath}
\usepackage{bbm}
\usepackage{dsfont}
\usepackage{pifont}
\usepackage{cite}
\usepackage{graphics}
\usepackage{graphicx}
\usepackage{epsfig}
\usepackage{subfigure}
\usepackage{amscd}
\usepackage{accents}
\usepackage{algorithm,algpseudocode}
\usepackage{multirow}

\begin{document}
	\title{Probabilistic Vehicle Trajectory Prediction over Occupancy Grid Map via Recurrent Neural Network}

		\author{\IEEEauthorblockN{ByeoungDo Kim, Chang Mook Kang, Jaekyum Kim, \\ Seung Hi Lee, Chung Choo Chung, and Jun Won Choi*}
		\IEEEauthorblockA{Hanyang University, Seoul, Korea\\
			Email: bdkim@spo.hanyang.ac.kr, kcm0728@hanyang.ac.kr,  jkkim@spo.hanyang.ac.kr, \\ shlee@ieee.org, cchung@hanyang.ac.kr and junwchoi@hanyang.ac.kr\\
*: corresponding author}
		
	}
	
	\maketitle
	
	\begin{abstract}
		In this paper, we propose an efficient vehicle trajectory prediction framework based on recurrent neural network.  Basically, the characteristic of the vehicle's trajectory is different from that of  regular moving objects since it is affected by various latent factors including road structure, traffic rules, and driver's intention. Previous state of the art approaches use sophisticated vehicle behavior model  describing these factors and derive the complex trajectory prediction algorithm, which requires a system designer to conduct intensive model optimization for practical use. Our approach is data-driven and simple to use in that it learns complex behavior of the vehicles from the massive amount of trajectory data through deep neural network model. The proposed trajectory prediction method employs the recurrent neural network called long short-term memory (LSTM) to analyze the temporal behavior and predict the future coordinate of the surrounding vehicles. The proposed scheme feeds the sequence of vehicles' coordinates obtained from sensor measurements to the LSTM and produces the probabilistic information on the future location of the vehicles over  occupancy grid map. The experiments conducted using the data collected from highway driving show that the proposed method can produce reasonably good estimate of future trajectory.
		%The proposed method estimates the future trajectory of the  vehicles surrounding the ego-vehicle in dynamic environments.
	\end{abstract}
	
	
	% \fi
	\IEEEpeerreviewmaketitle
	
	\section{Introduction}
	In order to build fully autonomous vehicle, it is necessary to guarantee high degree of safety even in uncertain and dynamically changing environments.
	%This paper addresses a problem of predicting behavior of traffic participants around the ego-vehicle for autonomous driving systems.
	In order to serve this goal, an autonomous vehicle should be able to anticipate what would happen to its environment in the  future and respond to the change appropriately in advance.
	However, the behavior of the traffic participants (e.g. the vehicles surrounding the ego-vehicle) is often hard to predict since it is affected by various latent factors such as driver's intention, traffic situations, road structure, and so on.
	The prediction based on vehicle's dynamics model is accurate only for  very near future and it does not match with true trajectory well for long term prediction (more than one second).
	In addition, it is hard to know the accurate maneuver control (e.g. steering and acceleration) of the other vehicles so that the trajectory prediction based on vehicle's dynamics model might not be effective in practical scenarios.
	In order to facilitate path planning and collision avoidance needed to realize fully autonomous driving, we need a simple but powerful prediction framework which can analyze  complex temporal dynamics of  the traffic participants well.
	
	In fact, a problem of tracking the trajectory of moving targets  has been actively studied  in computer vision and robotics fields. However, as mentioned,   predicting the motion of traffic participants is not as simple as object tracking due to complex dynamics of the vehicles in traffic.
	%including relative position of the vehicles, traffic rules, drivers' intention, lane/road structure, etc.
	%Specifically, each vehicle moves in accordance with its own goal but sometimes,  their motion is restricted by their interactions.
	%in a way that they keep the inter-vehicle distance to prevent accidents.  %In other words, the trajectory of the vehicles participating in traffic develops in their own way interacting with each other.
	%Fig. 1 illustrates this point.
	%vehicle's motion pattern for different traffic scenario. Without consideration of these factors, it is hard to achieve good accuracy in prediction.S
	So far,  various approaches have been proposed to analyze  the vehicles' motion \cite{survey}.
	Vehicle motion model such as kinematic or dynamic models has been used for trajectory prediction \cite{kalman,imm}. Kalman filter  has been widely used to perform prediction accounting for the uncertainty in vehicle model \cite{kalman}. In order to improve the prediction accuracy further, Bayesian filtering techniques such as
	the context-dependent interactive multiple model  filter \cite{imm} and Monte-Carlo method \cite{mc} have been proposed.
	%Bayeisan filtering approach is widely used to predict the trajectory of the vehicle.
	Recently,  machine learning techniques were employed to learn the complex model capturing driver's maneuver intention and their interactions from the data. The vehicle trajectory generating model has been learned through Gaussian process  \cite{gp,gp2} and Gaussian mixture model  \cite{gm}. More sophisticated models considering the vehicle interactions  were introduced, including dynamic Bayesian network \cite{dillmann} and coupled hidden Markov model \cite{cvpr}.   Though these methods yield better prediction accuracy, complexity for running training and inference algorithms on these models is significantly high. In addition, these models involve human's interpretation on the factors determining the trajectory so careful parameter tuning is needed for various environments.
	
	%Hence, dynamic behavior of traffic participants due to these factors should be captured in the model used for prediction.
	%When we predict the vehicle trajectory, it is also important to handle a certain degree of uncertainty using a probabilistic framework.
	%There are various source of uncertainties in the traffic model so that the predicted trajectory should be handled stochastically.
	%n essence, prediction of vehicle trajectory is a complex task, which needs sophisticated and dynamic traffic model.
	
	In this paper, we introduce a new efficient vehicle trajectory prediction framework based on the deep neural network.
	In recent years, the deep neural network has received much attention since it has showed great performance for a variety of machine learning tasks \cite{deep}. Among the several DNN architectures, recurrent neural network (RNN) is widely used to  learn temporal dynamics in the time-series data.   In particular, the  long short term memory (LSTM) architecture has been successfully applied to analyze the sequential structure underlying in text, speech, and financial data \cite{lstm}. The attractive feature of the deep neural network models is that using numerous data sources, they can find the features that are robust to a variety of changes in data.
	In our work, we employ LSTM to understand complex dynamics of vehicle motions. The proposed trajectory prediction system inputs the coordinates and velocities of the surrounding vehicles obtained from the sensor measurements to the LSTM and produces the vehicle's future location after $\Delta$ seconds. In order to train the LSTM, we use a long  record of the trajectory data acquired by long-term driving on real road and let the LSTM to predict the future coordinate based on the past trajectory input.
	%As a result, the proposed scheme  benefits from the abundance of the trajectory data that can be easily realized by recent rapid advance of cloud systems.
	In order to handle uncertainty in making prediction, the LSTM is designed to produce the probability of occupancy for the surrounding vehicles on the occupancy grid map.  The experiments conducted with the data collected from
	highway driving show that the proposed method offers better prediction accuracy over the existing Kalman filter-based method.

	Recently, we found that similar LSTM-based trajectory analysis has been proposed independently in \cite{trivedi,peter}. Note that our work is different from their methods in that while their focus is on classification of the vehicles' maneuvers at intersection and object tracking, the proposed method predicts the future  trajectory of the vehicles over the occupancy grid map.
	
	The rest of this paper is organized as follows. In Section II, we describe the system setup used for developing the proposed prediction framework.
	In Section III, we briefly introduce the structure of the LSTM and in Section IV, we describe the details on the proposed vehicle trajectory prediction framework.
	In Section V, the experimental results are provided and the paper is concluded in Section VI.
	%is not always correct, we assume that it is expressed as distribution function. In order to capture the interactions between multiple vehicles, we input the trajectory data of all vehicles discovered by the ego-vehicle. Using the vehicle's trajectory data, we train the LSTM such that it predicts the coordinates of the vehicle in the future. In order to represent the future location using probability, we adopt occupancy grid map which divides the region around the ego-vehicle into grid and indicates whether the grid is occupied by a car.  , for each grid element, the LSTM produces the probability that the target vehicle moved to the position indicated by grid. Finally, using actual space of the vehicle and distribution, we  Finally, we can combine the probability of each vehicle to get the probability of occupancy in traffic.
	
	%The proposed scheme has an advantage over the existing approaches.  Since feature is learned using data, we do not need complicated model.  Owing to availability of the communications and
	%data cloud, we can use ample amount of training data which cars send. We use only location not velocity or yaw is not needed.  The proposed method does not need labels that need human labor. Statistical description of location make more accurate assessment of prediction to a certain degree of confidence.
	
	\begin{figure}[t]
		\centering
		\includegraphics[scale=0.35]{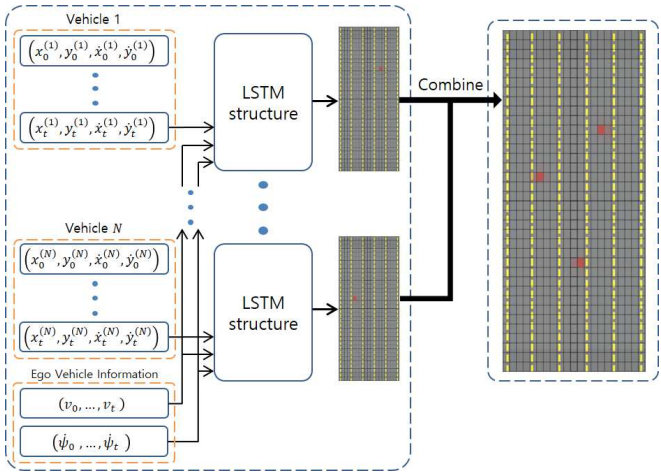}
		%\caption{Overview of the proposed system model}
		\caption{The structure of the proposed trajectory prediction (The coordinate $(x_t^{(i)},y_t^{(i)})$ and the velocity pair $(\dot{x}_t^{(i)},\dot{y}_t^{(i)})$ denote the position and velocity measurement of the surrounding vehicles and $\psi_t$ and $v_t$ denote the yaw rate and the velocity measurements of the ego-vehicle.)}
		\label{fig:system}
	\end{figure}
	
	\section{System Description} \label{sec_sys_model}
	
	In this section, we provide the basic system description for the proposed trajectory prediction method.
	
	\begin{figure}
		\centering
		\includegraphics[scale=0.2]{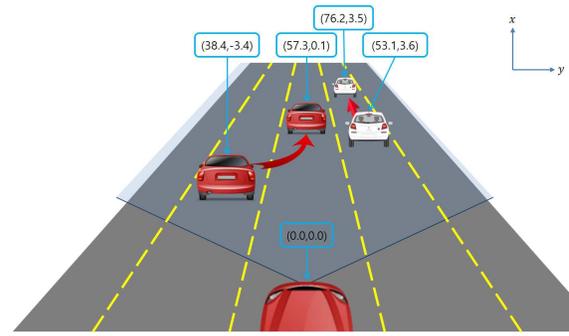}
		\caption{The relative coordinate $(x,y)$ of the vehicles to ego-vehicle}
		\label{fig:coord}
	\end{figure}
	
	\subsection{Vehicle Trajectory}
	Fig. \ref{fig:system} depicts the system model for the proposed scheme.
	%The coordinate $(x,y)$  of the vehicle represents the location of the vehicle center relative to the location of the ego-vehicle.
	The coordinate $(x,y)$ of the vehicle represents the location of the center of a target vehicle relative to that of the ego-vehicle as shown in Fig. \ref{fig:coord}. Note that $x$ and $y$ indicate the relative location in longitudinal and lateral directions, respectively.
	We let the coordinate of the ego-vehicle be $(0,0)$ and the coordinate of the $i$th surrounding vehicle at the time step $t$ is denoted as $(x_t^{(i)},y_t^{(t)})$.
	In addition, the velocity pair $(\dot{x}_t^{(i)},\dot{y}_t^{(i)})$ represents relative velocity of the $i$th vehicle.
	We assume that the coordinate information of the vehicle is obtained by processing the measurements from camera, radar, and Lidar sensors.
	%This information is subject to errors which will be accounted in vehicle trajectory projection step.
	%
	The trajectory of the $i$th vehicle is determined by the sequence of the coordinates $...,(x_{t-1}^{(i)},y_{t-1}^{(i)}), (x_{t}^{(i)},y_{t}^{(i)}), (x_{t+1}^{(i+1)},y_{t+1}^{(i+1)}), ...$, where each coordinate is acquired every $T_s$ second.
%	\begin{bfseries}
%	Since the information on the global velocity and heading angle of the vehicles can be inferred in the trajectory data, relative velocity data, and ego vehicle data, we do not use them explicitly for our trajectory prediction.
%	\end{bfseries}
	%We consider the coordinate of the surrounding vehicles in the range of -20 to 20 meters in lateral direction and 0 to 100 meters in the longitudinal direction.
	We consider the coordinate of the surrounding vehicles in the range of -9.2 to 9.2 meters in lateral direction and 0 to 180 meters in the longitudinal direction. Such range is determined by considering the valid detection range of the sensors and the highway environments in which the data is collected.
		
	%In order to capture the dynamic coupling and interactions between the vehicles, the proposed system uses the trajectory data for the $N$ nearest vehicles.
	%Note that the proposed scheme does not need complicated structure of traffic model which requires large efforts to optimize and tune model accuracy.
	
	\begin{figure}
		\centering
		\includegraphics[scale=0.2]{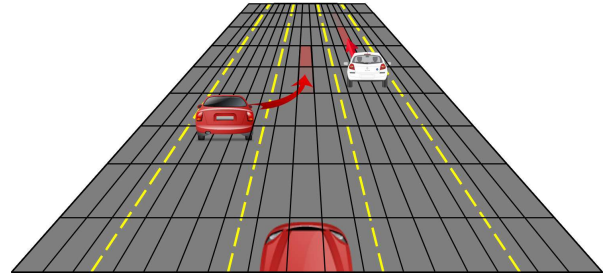}
		\caption{The occupancy grid map used by the proposed scheme}
		\label{fig:ogm}
	\end{figure}
	
	\subsection{Occupancy Grid Map}
	
	In essence, at the current time step $t$, the objective of the trajectory prediction is to estimate the future coordinate $(x_{t+\Delta}^{(i)},y_{t+\Delta}^{(t)})$ based on the past trajectory data $...,(x_{t-1}^{(i)},y_{t-1}^{(i)},\dot{x}_{t-1}^{(i)},\dot{y}_{t-1}^{(i)}),(x_{t}^{(i)},y_{t}^{(i)},\dot{x}_{t}^{(i)},\dot{y}_{t}^{(i)})$.
	The proposed system produces the prediction results in the form of probability using the occupancy grid map.
	The occupancy grid map is widely adopted for probabilistic localization and mapping in robotics \cite{thron}. We use the occupancy grid map to reflect the uncertainty of the predicted trajectory with probability.
	%The proposed trajectory prediction method outputs the probability that each grid element is occupied by at least one of all surrounding vehicles.
	The occupancy grid map is constructed by partitioning the range under consideration into $M_{x} \times M_{y}$ grid elements (see Fig.~\ref{fig:ogm}).
	We determine the grid size such that the grid element approximately covers the quarter lane to recognize the movement of the vehicle on same lane as well as length of the vehicle.
		Note that the occupancy grid map is aligned with heading angle of ego-vehicle since the coordinates of the vehicles are obtained through the sensors of the ego-vehicle.
	
	\section{Structure of LSTM}
	
	\begin{figure}
		\centering
		\includegraphics[scale=0.5]{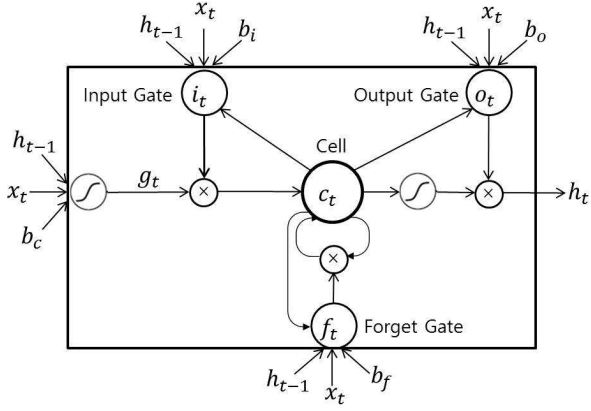}
		\caption{Basic LSTM structure}
		\label{fig:lstm}
	\end{figure}
	
	In this section, we briefly introduce the structure of the LSTM.
	Fig. \ref{fig:lstm} depicts the structure of the LSTM.
	The LSTM was developed to overcome the shortcoming of RNN, i.e., vanishing and exploding gradient problem \cite{lstm}.
	The LSTM has a memory called ``cell" which is used to store the state vector summarizing the sequence of the past input data.
	The current state of the cell is updated according to the input, the output, and the previous state of the cell. Unlike recurrent neural network, LSTM has a gating control mechanism that allows the network to
	forget past state in the memory or learn when to update its state given new information. Let $c_t$ be the state of the memory cell at the current time step $t$. Then, $c_t$ is
	updated by the following recursive equations
	\begin{align}
	i_t &= \sigma(W_{xi} x_t + W_{hi}h_{t-1} + b_i)   \\
	f_t &= \sigma(W_{xf} x_t + W_{hf}h_{t-1} + b_f)   \\
	o_t &= \sigma(W_{xo} x_t + W_{ho}h_{t-1} + b_o)  \\
	g_t &= \tanh(W_{xc} x_t + W_{hc}h_{t-1} + b_c)   \\
	c_t &= f_t \odot c_{t-1} + i_t \odot g_t \\
	h_t &= o_t \odot \tanh(c_t)
	\end{align}
	where
	\begin{itemize}
		\item $\sigma(x) = \frac{1}{1+e^{-x}}$: sigmoid function
		\item $x \odot y$: element-wise product
		\item $W_{xi}, W_{hi}, W_{xf}, W_{hf}, W_{xo}, W_{ho},W_{xc},W_{hc}$: weight matrix for linear transformation
		\item $b_i, b_f, b_o, b_c$: bias vector
		\item $i_t$: input gating vector
		\item $f_t$: forget gating vector
		\item $o_t$: output gating vector
		\item $g_t$: state update vector
		\item $h_t$:  output hidden state vector
	\end{itemize}
	Note that with the configuration $f_t = [0,...,0]$, the network can forget the information $c_{t-1}$ stored in the memory cell. The input gate $i_t$ and the output gate $f_t$ can control  the information flow from the input and to the output, respectively. Note that the behavior of the gate control is learned from data as well.
	%The gating mechanism allows the LSTM to learn complex and long-term dynamics.
	When unfolded in time, the LSTM is considered to be a deep model. We can also make the cell to have deep architecture by adding additional layers into the cell.
	In order to extract the information relevant to the given task, we add additional output network to the hidden state $h_t$.
	%For example, for multi-class classification task, we add softmax layer which contains the linear transformation of $h_t$ followed by
	For instance, for multi-class classification task, we use a softmax layer which contains the linear transformation of $h_t$ followed by
	the softmax function $\frac{\exp(a_i)}{\exp(\sum_j a_j)}$.
	
	\section{Proposed Vehicle Trajectory Prediction}
	
	In this section, we describe the proposed trajectory prediction framework in details.
	
	\subsection{Proposed Trajectory Prediction Technique}
	Fig. \ref{fig:system} depicts the structure of the proposed system.
	The ego-vehicle estimates the current coordinates and velocities of the $N$ surrounding vehicles  and feeds them into the $N$ LSTMs for each.
	Each of these $N$ LSTMs produces the predicted result corresponding to each of the $N$ nearest vehicles. Note that the same network parameters  are used for all $N$ LSTMs.
	%Since the velocity of the vehicles can be inferred from the coordinate information, we do not feed the velocity information into the LSTMs.
	%The LSTM is trained to produce the estimate of future coordinate $(x_{t+\Delta}^{(i)},y_{t+\Delta}^{(i)})$ given the past trajectory $..., (x_{t-1}^{(i)},y_{t-1}^{(i)}), (x_t^{(i)},y_t^{(i)})$.
	%The LSTM is trained to produce the probability of occupation for each grid element for the time step $t+\Delta$ given the past trajectory data $..., (x_{t-1}^{(i)},y_{t-1}^{(i)}), (x_t^{(i)},y_t^{(i)})$.
	Since the trajectory information is obtained from the sensors of the ego-vehicle, the relative position of the surrounding vehicles would change according to the motion of the ego-vehicle. To compensate such coordinate change, we input the yaw rate $\psi_t$ and the velocity $v_t$ of the ego-vehicle to the LSTM along with the vehicle coordinates (see Fig.~\ref{fig:system}). Note that the yaw rate and the velocity of the ego-vehicle are measured using  inertial measurement unit (IMU) sensor.
	
	It might be necessary to make prediction with different terms  e.g., $\Delta= 0.5s, 1.0s$, and $2.0s$. In this case, we train the LSTM independently for different tasks  $\Delta= 0.5s, 1.0s$, and $2.0s$, thus yielding the multiple sets of the LSTM parameters. Once trained, we can load the parameter set corresponding to one of multiple future time steps.
	 The LSTM produces the probability of occupancy for each grid element of the occupancy grid map. Letting $(i_x,i_y)$ be the two dimensional index for the occupancy grid, the softmax layer in the $i$th LSTM produces the probability $P_o^{(i)}(i_x,i_y)$ for the grid element $(i_x,i_y)$. Finally, we combine the outputs of the $N$ LSTMs using
	\begin{align}
	P_o(i_x,i_y) = 1- \prod_{i=1}^{N} (1-P_o^{(i)}(i_x,i_y))
	\end{align}
	Note that the probability of occupancy $P_o(i_x,i_y)$ summarizes the prediction of the future trajectory for all $N$ vehicles in the single map. It provides the comprehensive  view on how $N$ surrounding traffic participants would behave after $\Delta$ seconds.

Alternatively, we can consider the different type of prediction system that directly produces the predicted coordinate of the vehicles. This system is useful when we do not need probabilistic information on the trajectory but the deterministic predicted value of the vehicle's future coordinate. In this case, we can train the same LSTM architecture using the loss function defined under the regression task. Instead of using the softmax layer, the system directly produces the two real coordinate values in x axis and y axis.
	
	\subsection{Training of LSTM}
	Using the vehicle equipped with sensors,  the trajectory data for the surrounding vehicles can be collected  through long-term driving on real road. The coordinate of the vehicles can be generated by employing the localization algorithm based on sensor fusion. From the trajectory history for all $N$ nearest vehicles, we extract the data for individual vehicle and combine them to generate the training data. The training data containing the trajectory of each individual vehicle is used to train the single LSTM. %Our trajectory prediction problem is formulated into the multi-class classification problem where we choose one grid element occupied by the target vehicle in the future among all $M$ grid elements.
	We formulate the trajectory prediction problem as a multi-class classification problem where one grid element occupied by the target vehicle should be chosen among all $M$ grid elements.
	Given the sequence of the coordinates $..., (x_{t-1}^{(i)},y_{t-1}^{(i)}), (x_t^{(i)},y_t^{(i)})$ for the $i$th vehicle, the label is automatically generated from the coordinate of the vehicle $\Delta$ seconds later. We use one hot encoding to generate the label. For example, the label for the $t$th training example $o_t = [\underbrace{0,...,0}_{l-1 \mbox{ times}},1,0,...,0]$  indicates the occupancy of the $l$th element in the occupancy grid map.
 Since we use the coordinate after $\Delta$ seconds as a label, we do not need human labor to label the data. Thus, the proposed system benefits from automatic logging and supports online system that performs training in normal driving. In order to train the LSTM, we minimize the negative log-likelihood function
	\begin{align}
	L (\mathbf{w})=&  -\sum_{t=1}^{J} \sum_{m=1}^{M} o_{t,m} \ln z_{t,m} + (1-o_{t,m}) \ln (1-z_{t,m}), \\
& + \lambda \Omega(\mathbf{w})
	\end{align}
	where $\mathbf{w}$ is the parameter of the neural network, $J$ is the total number of the training examples, $o_{t,m}$ is the $m$th entry of $o_{t}$, $z_{t,m}$ is the
	$m$th output of the softmax layer associated with the label $o_{t,m}$, and $\Omega(\mathbf{w})$ is the regularization term with the parameter $\lambda$.
	In order to optimize the network parameters of the LSTM, we employ ``back propagation through time" (BPTT) algorithm with mini-batch size $B$ \cite{lstm}.

The deterministic prediction system that produces the future coordinate directly uses the following loss function
	\begin{align}
	L' (\mathbf{w})=&  \sum_{t=1}^{J} \frac{1}{2} \left( |x_{t+\Delta} - z_{t,x}|^2 + |y_{t+\Delta} - z_{t,y}|^2\right), \\
& + \lambda \Omega(\mathbf{w})
	\end{align}
where $z_{t,x}$ and $z_{t,y}$ are the predicted values of the coordinate produced by the neural network for the $t$th example given. The rest of the training procedure is similar to that for the occupancy grid map-based prediction.

	%\subsection{Post-processing }
	%Since the vehicle coordinate obtained from the sensor measurement is subject to error, it is necessary to take it into account to produce the predicted results on the occupancy grid map. As shown in the Fig.~x., the accuracy of position estimation would change according to the location of the surrounding vehicle. The greater the distance from the vehicle, the less accurate the measured value. If we divide the occupancy grid map into $P$ regions, we can empirically obtain the distribution of position estimate with respect to the true position. At the $p$th region, we model the position estimate as $(x_t^{(i,true)},y_t^{(i,true)})=(x_t^{(i)},y_t^{(i)})+(e_x^{(p)},e_y^{(p)})$, where the distribution $f(e_x^{(p)},e_y^{(p)})$ is known.  Since the proposed scheme does not take that into account for prediction. Hence,
	
	\section{Experiments}
	
	\subsection{Experiment Description}
	
	\begin{figure}
		\centering
		\subfigure[Test vehicle] {\includegraphics[scale=0.35]{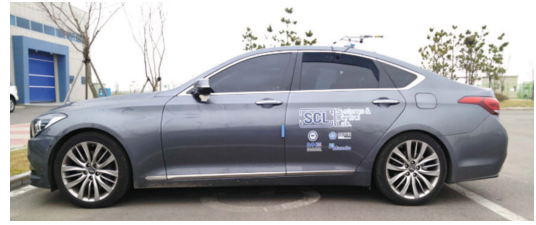}}\\
		\subfigure[Overall structure of data collection]{\includegraphics[scale=0.35]{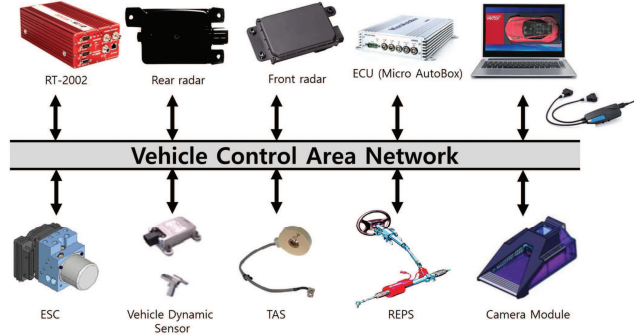}}\\
		\caption{Description of test vehicle}
		\label{fig:vehicle}
	\end{figure}
	
	 In our experiment, we collected the training data from long hours of highway driving in the suburbs of Seoul, Korea.
	As a test vehicle, we use a Hyundai Genesis equipped with a long range radar from Delphi and a single front camera from Mobileye.
	Fig.~\ref{fig:vehicle} (a) and (b) show the picture of the test vehicle and the system configurations for data logging, respectively.
	During driving, the test vehicle collects the sensor measurements which are processed to obtain the relative coordinates of the surrounding vehicles. The yaw rate and velocity of the ego vehicle is also logged in the system. The sample rate for the coordinate data is set to 10 ms.
	However, we found that the data logged every 10 ms is subject to large noise and  often cut off due to asynchronous sampling timing. In order to resolve these issues, we synchronized the data by averaging the coordinate samples for 100 ms of duration. As a result, new trajectory input data is updated every 100 ms.
	Acquisition of the training data was conducted under various real driving situations on highway such as cruising, lane change and merge junction. The collected data was 3730 seconds total from 26 scenarios.
	Of the collected 3730 seconds of data, there are a total of 1325 vehicles detected for more than 4 seconds within the valid detection range.  Among 1325 cases, 1126 cases (85\%) were used for training and 199 cases (15\%) were used for validation.
	%In order to train the LSTM, we extracted many 4 second trajectory sequences where the vehicle coordinate and the yaw rate are logged every 100 ms.
	%The given 4 second data is divided into input data and target data according to the prediction scenario. For example, in the case of a scenario that predicts 1.0 second later, the interval from 0 to 3 seconds is used as the input data and the interval between 1 second and 4 seconds is used as the target data.
	
	%\begin{table}[]
	%	\centering
	%	\caption{experimental hyperparameter setup}
	%	\label{setup}
	%	\begin{tabular}{|c|c|c|c|c|}
	%		\hline
	%		\begin{tabular}[c]{@{}c@{}}batch\\ size\end{tabular} & \begin{tabular}[c]{@{}c@{}}learning\\ rate\end{tabular} & \begin{tabular}[c]{@{}c@{}}decreasing\\ factor\end{tabular} & \begin{tabular}[c]{@{}c@{}}cost\\ function\end{tabular} & optimizer \\ \hline
	%		\multicolumn{1}{|r|}{24} & \multicolumn{1}{r|}{1.0} & MAE & \begin{tabular}[c]{@{}c@{}}negative\\ log-likelihood\end{tabular} & \begin{tabular}[c]{@{}c@{}}stochastic\\ gradient descent\end{tabular} \\ \hline
	%	\end{tabular}
	%\end{table}
	
	\begin{figure}[h]
		\centering
		\includegraphics[scale=0.4]{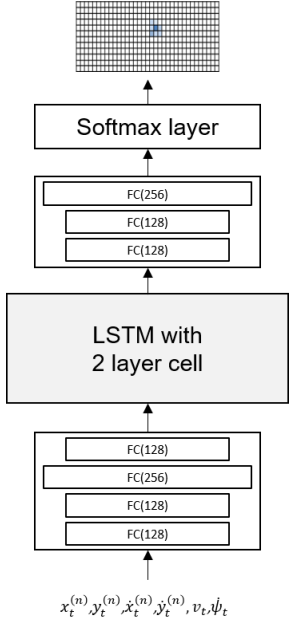}
%		\subfigure[]{\includegraphics[scale=0.25]{structure_05_00}\label{fig:struc_05}}
%		\hspace{0.5cm}
%		\subfigure[]{\includegraphics[scale=0.25]{structure_10_00}\label{fig:struc_10}}
%		\hspace{0.5cm}
%		\subfigure[]{\includegraphics[scale=0.25]{structure_20_00}\label{fig:struc_20}}
		\caption{The structure of LSTM used for the proposed method}
		\label{fig:structure}
	\end{figure}
	
	\subsection{System Configurations}
	
	In our system, we consider the occupancy grid map with size $(M_x, M_y) = (36,21)$ (i.e., 36 elements in longitudinal direction and 21 elements in lateral direction). Each grid element occupies an area of the width 0.875 meter and height 5 meter.
	When a target vehicle gets out of the boundary of the occupancy grid map, we label it as ``out of boundary" class.
	%According to the size of the occupancy grid map, the number of output labels in the softmax layer is 198.
	%However since we add one label considering whether there is a cutoff between the collected data or the target object is outside the boundary, the size of the output of the neural network is 199.
	In our experiments, we consider three scenarios with $\Delta =  0.5 s, 1.0s, $ and $2.0s$.
	We empirically search for the best performing architecture of the LSTM for each scenario.
		The architecture of the proposed system consists of the concatenation of the input fully-connected layers, the LSTM with two layer cell memory, the output fully-connected layers and one softmax layer (see  Fig. \ref{fig:structure}).  The input and output fully-connected layers lead to higher capacity to fit the complex structure of vehicle trajectory and we find that they offer  the non-negligible improvement in prediction accuracy.
%The LSTM layers extract time-sequential information and the final states of the LSTM layers are processed to probabilistic occupancy grid map through followed multiple fully-connected layers and softmax layer.
The numbers of layers and nodes were chosen with grid search and the number of nodes at the softmax layer was decided  by the size of occupancy grid map plus out of boundary class, i.e.,  $757 (=M_x M_y + 1)$.
	We set the initial learning rate to 0.001 and gradually decreased it whenever the validation error stops to improve. The mini-batch size $B$ was set to 40.  $\L_2$ regularization term was used only for the weights of the fully-connected layers and the softmax layer. The regularization parameter $\lambda$ was set to 0.0005.
	
	\begin{table}[h]
	\centering
	\caption{Prediction accuracy  of the proposed scheme and Kalman filter}
	\label{result_ogm}
	\begin{tabular}{|c|r|r|r|r|}
	\hline
	 & \multicolumn{1}{c|}{\begin{tabular}[c]{@{}c@{}}Prediction term \\ $\Delta$\end{tabular}} &
	\multicolumn{1}{c|}{\begin{tabular}[c]{@{}c@{}}MAE X\\ (grid)\end{tabular}} & \multicolumn{1}{c|}{\begin{tabular}[c]{@{}c@{}}MAE Y\\ (grid)\end{tabular}} & \multicolumn{1}{c|}{\begin{tabular}[c]{@{}c@{}}MAE\\ (grid)\end{tabular}} \\ \hline
	\multirow{3}{*}{\begin{tabular}[c]{@{}c@{}}Proposed\\ Scheme \end{tabular}}
	& 0.5 & 0.29 & 0.52 & 0.77 \\ \cline{2-5}
	& 1.0 & 0.27 & 0.70 & 0.88 \\ \cline{2-5}
	& 2.0 & 0.44 & 1.06 & 1.31 \\ \hline
	\multirow{3}{*}{\begin{tabular}[c]{@{}c@{}}Kalman\\ Filter\end{tabular}}
	& 0.5 & 0.51 & 1.55 & 1.73 \\ \cline{2-5}
	& 1.0 & 0.96 & 2.99 & 3.26 \\ \cline{2-5}
	& 2.0 & 2.07 & 5.84 & 6.36 \\ \hline
	\end{tabular}
	\end{table}
	
	\begin{table}[h]
	\centering
	\caption{Prediction accuracy of the proposed scheme trained for regression task}
	\label{result_reg}
	\begin{tabular}{|c|r|r|r|r|}
	\hline
	& \multicolumn{1}{c|}{\begin{tabular}[c]{@{}c@{}}Prediction term \\ $\Delta$\end{tabular}} &
	\multicolumn{1}{c|}{\begin{tabular}[c]{@{}c@{}}MAE X\\ (m)\end{tabular}} & \multicolumn{1}{c|}{\begin{tabular}[c]{@{}c@{}}MAE Y\\ (m)\end{tabular}} & \multicolumn{1}{c|}{\begin{tabular}[c]{@{}c@{}}MAE\\ (m)\end{tabular}} \\ \hline
	\multirow{3}{*}{\begin{tabular}[c]{@{}c@{}}Prediction\\ Scheme \end{tabular}}
	& 0.5 & 0.41 & 0.29 & 0.59 \\ \cline{2-5}
	& 1.0 & 0.63 & 0.46 & 0.88 \\ \cline{2-5}
	& 2.0 & 1.15 & 0.69 & 1.51 \\ \hline
	\multirow{3}{*}{\begin{tabular}[c]{@{}c@{}}Kalman\\ Filter\end{tabular}}
	& 0.5 & 2.57 & 1.37 & 3.12 \\ \cline{2-5}
	& 1.0 & 4.81 & 2.61 & 5.77 \\ \cline{2-5}
	& 2.0 & 10.32 & 5.11 & 11.92 \\ \hline
	\end{tabular}
	\end{table}
	
	\subsection{Experimental Results}
	
	For performance evaluation in classification task, top-$K$ classification errors are widely used. Since our objective is to predict the coordinates of surrounding vehicles, we need an appropriate performance metric for performance evaluation. Since the proposed system produces the probability of occupancy, we need an appropriate performance metric. In our experiments, we use the weighted mean absolute error (MAE) as a performance metric.
	\begin{align} \label{eq:mae}
	MAE= \frac{1}{N}\sum_{i=1}^{N}\sum_{i_x=1}^{M_{x}}\sum_{i_y=1}^{M_{y}}{\left\| \begin{bmatrix}i_x \\ i_y \end{bmatrix} - \begin{bmatrix} i_x^{(desired)} \\ i_y^{(desired)} \end{bmatrix} \right\| P_o^{(i)}(i_x,i_y)}
	\end{align}
	where $(i_x^{(desired)}, i_y^{(desired)})$ is the index of the desired grid element in the label  and $P_o^{(i)}(i_x,i_y)$ is the probability occupancy for the grid element $(i_x,i_y)$.
		
	Table \ref{result_ogm} shows the accuracy of the trajectory prediction for the proposed algorithm. Besides the MAE defined in (\ref{eq:mae}), we provide the MAE in longitudinal direction (MAE X) and that in lateral direction (MAE Y).
	%Both top-1 and top-9 errors are provided as well.
		Obviously, the prediction accuracy of the proposed scheme is better for the case of short-term prediction as compared to long-term prediction. Note that for $\Delta = 0.5s$ and $1.0s$, the MAE metric is smaller than 1.0 grid. Though the MAE can go up to 1.5 grid for $\Delta = 2.0s$, such MAE levels seem to be good enough for the purpose of risk assessment and path planning.
	We also compare the performance of the proposed scheme with that of the Kalman filter-based prediction. As shown in Table \ref{result_ogm} and \ref{result_reg}, the LSTM outperforms the conventional Kalman filter for all cases considered. Note that the performance gap between them gets larger for long-term prediction. This shows that the proposed LSTM-based method is good at analyzing the complex patterns in vehicle motion underlying in long-term prediction.
	We observe that the MAE Y of the Kalman filter is much larger than that of the proposed scheme. This means that the proposed method is more effective in predicting the lateral motion than the Kalman filter.
 From the results,	we also notice that while the Kalman filter works well for the nominal cases where the trajectory of the vehicle looks smooth and predictable, it exhibits significant error for some difficult and ambiguous cases. On the contrary, the proposed scheme maintains reasonably good prediction accuracy for various challenging cases.
	
		We also test the deterministic prediction system trained under the regression task.  In Table \ref{result_reg}, the MAEs achieved by the proposed method is compared with that of the Kalman filter. Note that the proposed scheme achieves the significant gain in prediction accuracy over the Kalman filter. This shows the proposed regression system is also good at predicting the future coordinate of the vehicles.

	%We also compare the performance of the proposed scheme with that of the Kalman filter-based prediction. As shown in Table \ref{result}, the LSTM outperforms the conventional Kalman filter in terms of MAE metric. We notice that while the Kalman filter works well for the nominal cases where the trajectory of the vehicle looks smooth and easily predictable, it exhibits significant error for some difficult cases. On the other hand, the proposed scheme yields reasonably good performance for various cases.
	
	   \begin{figure}[h]
		\centering
		\subfigure[]{\includegraphics[scale=0.12]{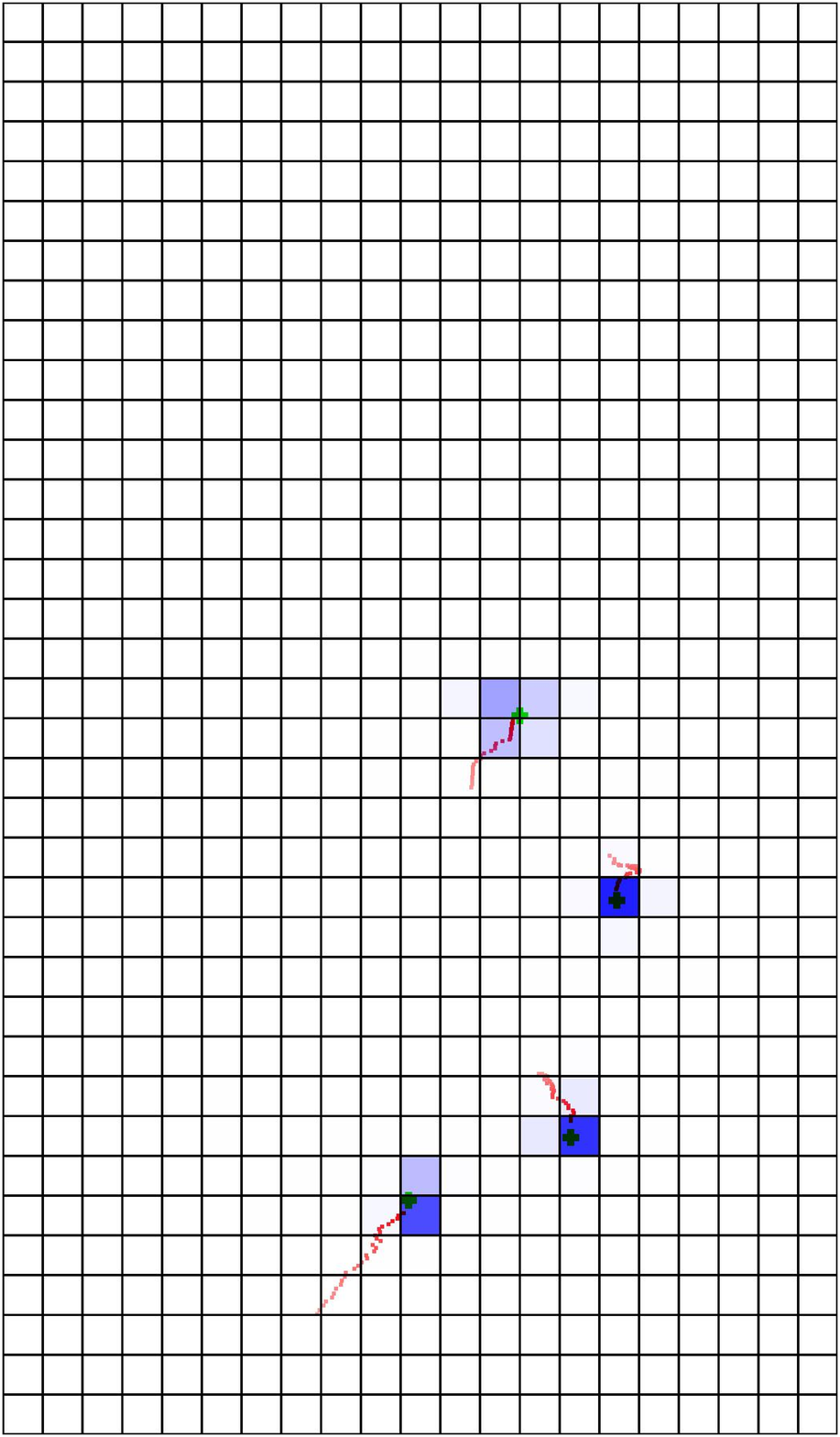}\label{fig:res_05}}
		\hspace{0.5cm}
		\subfigure[]{\includegraphics[scale=0.12]{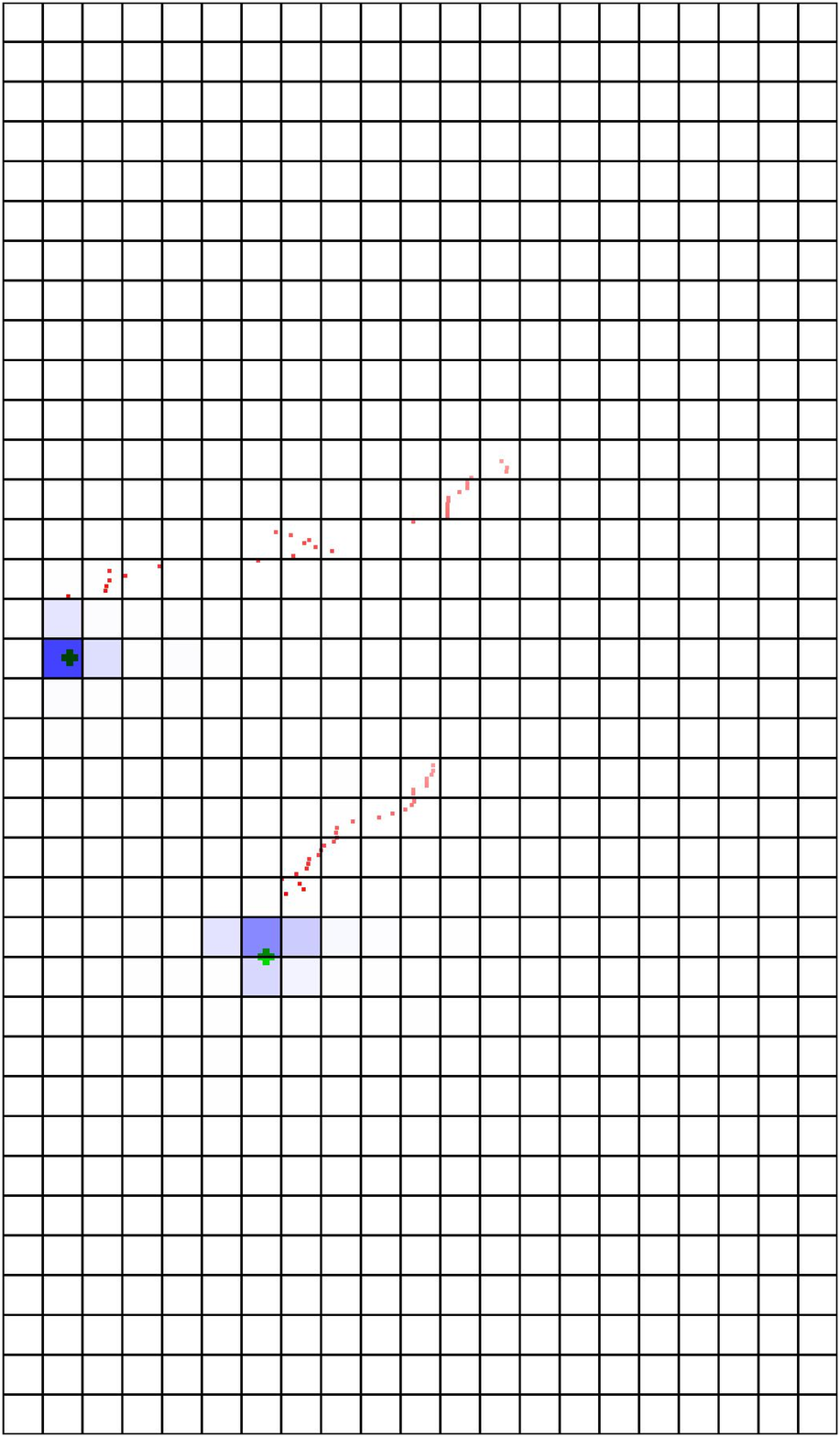}\label{fig:res_10}}
		\hspace{0.5cm}
		\subfigure[]{\includegraphics[scale=0.12]{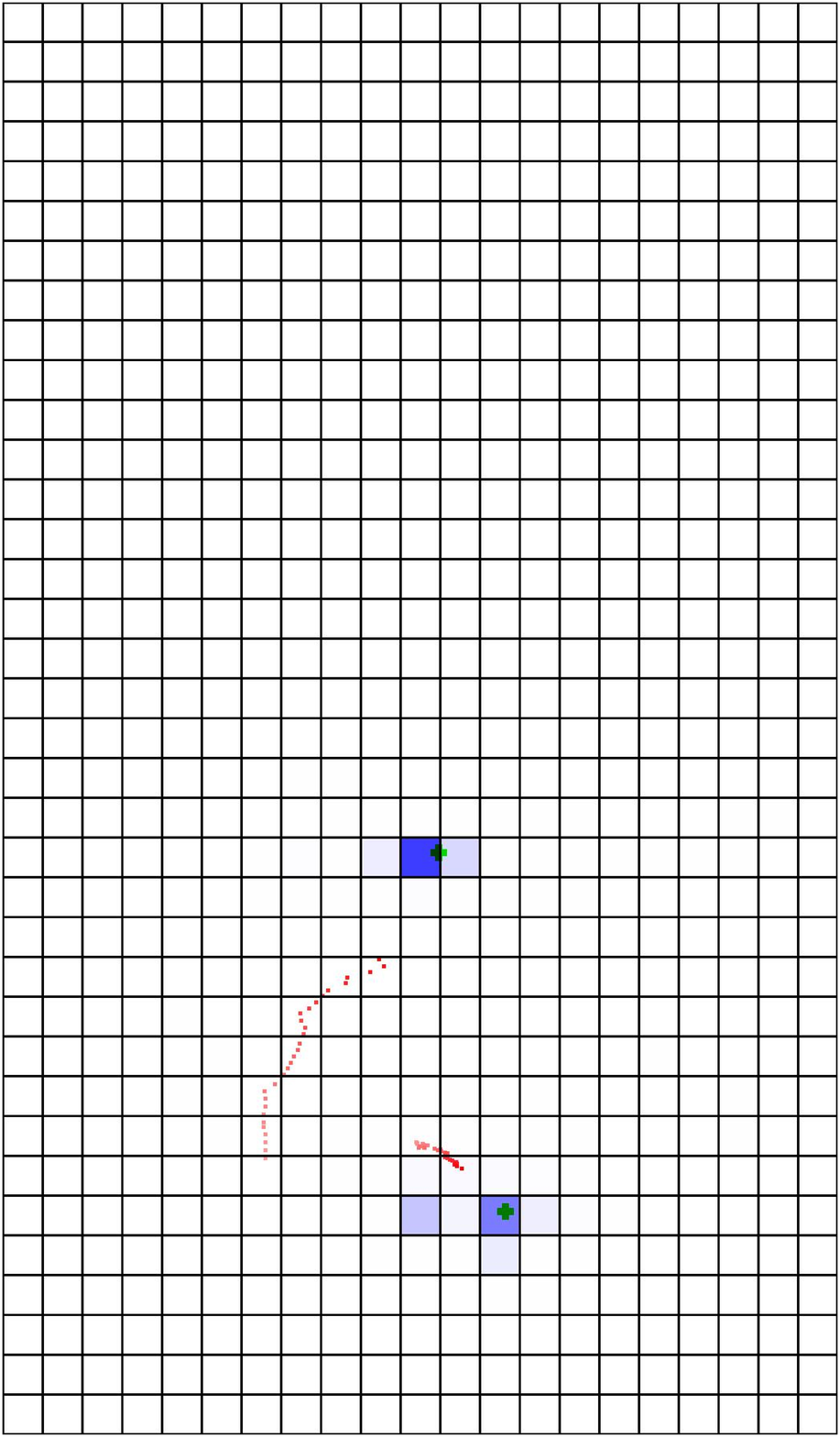}\label{fig:res_20}}
		%\caption{The results of the prediction for the case of (a) $\Delta=0.5s$, (b) $\Delta=1.0s$, and (c) $\Delta=2.0s$.}
		\caption{Several experimental results for the case of (a) $\Delta=0.5s$, (b) $\Delta=1.0s$, and (c) $\Delta=2.0s$.}
		\label{fig:result}
	\end{figure}
	
	Fig. \ref{fig:result} (a), (b), and (c) illustrate how the proposed algorithm predicts the trajectory for $\Delta=0.5s$, $\Delta=1.0s$, and  $\Delta=2.0s$, respectively. The small dots colored in red represent the past trajectory of the surrounding vehicles. The darker red color indicates more recently acquired coordinates. The cross mark colored with green indicates the future coordinate of the vehicle after $\Delta$ seconds, which should be predicted by the proposed method. The probability of the occupancy, which is the output of the proposed scheme, is shown on the occupancy grid map. Note that higher magnitude of the probability is translated into darker blue color.  %In  Fig. \ref{fig:result} (a), the three front vehicles are detected and one of them is trying to change the lane. blar blar blar.....
		Fig. \ref{fig:res_05} shows that the proposed system predicts the trajectory of four surrounding vehicles at good accuracy.
%Note that the prediction of the front-line vehicle placed at the point of intersection of grids shows distributed result on the surrounding grids of the intersection.
	The task of prediction seems more challenging in Fig. \ref{fig:res_10} and Fig. \ref{fig:res_20}. The surrounding vehicles exhibit long past trajectory and their motion patterns look a bit complex. In Fig. \ref{fig:res_10}, two front vehicles are trying lane changes. We see that the proposed method successfully predict the motion change of both vehicles after 1 second.
	Fig. \ref{fig:res_20} shows that the proposed method works reasonably well even for the challenging case $\Delta=2.0s$.  This shows that our method can be a promising solution to predict the behavior of traffic participants in real road and thus enhance  safety level for realizing fully autonomous driving.
	
	%We further compare the performance of the proposed algorithm to the conventional algorithms such as Kalman filter and particle filter. On this comparison, we only consider a few scenarios of the test data. Fig. x shows the results from the same environment, same time data of the conventional methods and the proposed algorithm. Fig.a shows the result of the Kalman filter using only the x, y coordinates and the speed of the target vehicle in an environment assuming a constant velocity model. Fig.b show the results of the particle filter. Fig.c shows the result of the proposed algorithm using only the x, y coordinates of the target vehicle and yaw rate of the ego vehicle.
		
	\section{Conclusions} \label{sec_conclusion}
	In this paper, we proposed the probabilistic vehicle trajectory prediction method based on the LSTM. Using the large amount of the trajectory data acquired by processing the sensor measurements during long-term driving, we trained the LSTM to learn complex dynamics of the surrounding vehicle's motion and predict its location in the future. 
%We used the occupancy grid map to represent the occupancy using probability and such probabilistic approach can benefit more reliable  path planning and risk assessment.
Our experiments performed using the data collected on real highway show that the proposed method provides the reasonably accurate prediction on the surrounding vehicles' trajectory.
	
\section*{Acknowledgment}
	
This work was supported by Institute for Information \& communications Technology Promotion (IITP) grant funded by the Korea government (MSIP) (No.20160001700022004).

\end{document}